\documentclass[11pt]{article}
\usepackage{acl2013}
\usepackage[usenames,dvipsnames]{xcolor}

\usepackage{times}
\usepackage{url}
\usepackage{alltt}
\usepackage{dblfloatfix}

\usepackage{drs}
\renewcommand\drscondfont[1]{\sffamily
\color{Bittersweet} #1}

\usepackage{textcomp}
\usepackage{latexsym}
\usepackage{float}
\usepackage{amsmath}
\usepackage{times}
\usepackage{url}
\usepackage{alltt}
\usepackage{latexsym}

\usepackage{tikz}
\usepackage{graphicx}
\usetikzlibrary{calc,trees,positioning,arrows,chains,shapes.geometric,%
    decorations.pathreplacing,decorations.pathmorphing,shapes,%
    matrix,shapes.symbols}

\tikzset{
>=stealth',
  punktchain/.style={
    rectangle, 
    rounded corners, 
    draw=black, very thick,
    text width=10em, 
    minimum height=3em, 
    text centered, 
    on chain},
  line/.style={draw, thick, <-},
  element/.style={
    tape,
    top color=white,
    bottom color=blue!50!black!60!,
    minimum width=8em,
    draw=blue!40!black!90, very thick,
    text width=10em, 
    minimum height=3.5em, 
    text centered, 
    on chain},
      punktempty/.style={
    rectangle, 
    rounded corners, 
    draw=white, very thick,
    text width=10em, 
    minimum height=3em, 
    text centered, 
    on chain},
  every join/.style={->, thick,shorten >=1pt},
  decoration={brace},
  tuborg/.style={decorate},
  tubnode/.style={midway, right=2pt},
}
\title{Towards Structural Natural Language Formalization: \\Mapping Discourse to Controlled Natural Language}
\author{Nicholas H. Kirk \\
  Computer Science Department \\ 
  Technische Universit\"{a}t M\"{u}nchen \\
  {\tt nicholas.kirk@tum.de} \\
}  

\begin{document}

\maketitle

\begin{abstract}
The author describes a conceptual study towards 
mapping grounded natural language discourse representation 
structures to instances of controlled language statements. 
This can be achieved via a pipeline of preexisting state of the art
technologies, namely natural language syntax to semantic discourse
mapping, and a reduction of the latter to controlled language discourse, 
given a set of previously learnt reduction rules.
Concludingly a description on evaluation, potential and limitations for ontology-based reasoning is presented.

\end{abstract}

\begin{figure*}
\scalebox{0.68}{
\begin{tikzpicture}
  [node distance=.8cm,
  start chain=going right,]
     \node[punktchain, join] (source) {\large Source text normalization (C1)};
     \node[punktchain, join] (toDRSNL) {\large text to $DRS_{NL}$ (C2)};
     \node[punktchain, join] (perfekt) {$ \substack{\color{black} \left\lbrace{\substack{\vspace{2mm}\text{\normalsize  colloquialism}\\ \vspace{2mm}\text{\normalsize  jargon}\\ \vspace{3mm}\text{\normalsize workaround}\\ \vspace{3mm}\text{\large ambiguous} \\ $\dots$}} \right\rbrace\vspace{3mm}\\ \text{\large classification (C3)}}$};
     \node[punktchain, join] () { {$DRS_{NL}$ to $DRS_{ CNL}$ \\Manipulation Engine (C4)}};
     \node[punktempty, join, ] (emperi) { $\substack{\color{black} \substack{\left\lbrace{\substack{\vspace{2mm}\text{\normalsize prove}\\\vspace{2mm} \text{\normalsize  paraphrase}\\ \vspace{2mm}\text{\normalsize  reason on}\\ \vspace{2mm} \text{\normalsize  store} \\ \vspace{2mm} \text{\normalsize  $\dots$}}} \color{white}\right\rbrace}\\ {\vspace{10mm} \color{black}\text{\large CNL statement}}} $};
  \end{tikzpicture}
  }
  \caption{Representation of an abstract structure-level only NL to CNL manipulator}
\end{figure*}
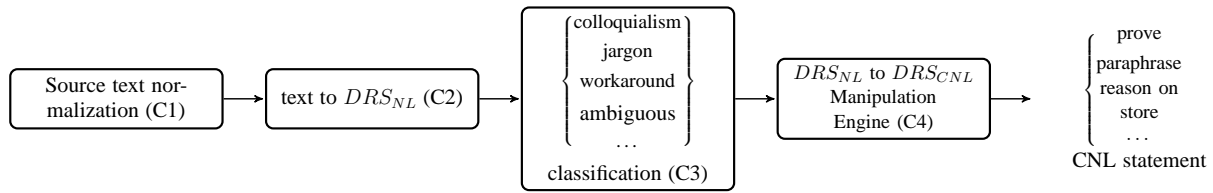

\section{Motivation}

Work towards the formalization of natural language has been pursued on both 
syntactic and semantic levels. Controlled Natural Languages (CNL) for instance provide 
an unambiguous set of syntactic rules and a controlled vocabulary \cite{wyner2010controlled},
while sharing human intelligibility with the original Natural Language (NL) from
which it derives \cite{kuhn2013understandability}. 
Approaches to pure semantic formalization have been done via symbolic and distributional characterizations \cite{blackburn2001inference,harris1981distributional}, to various 
extents of compositionality \cite{clarke2012context}.

An important and structural approach towards formalization of discourse is
Discourse Representation Theory (DRT) \cite{kamp1981theory,kamp1993discourse}, 
which makes use of inter- and intra-sentence discourse referents for anaphoric 
referencing and meaning preservation, and a set of semantic-level constraints over them. 
DRT maintains transformations to and from logic formalisms \cite{kamp1993discourse}, 
and has direct applications within the automated sentence construction domain
\cite{guenthner1984automatic,ifi-2010.0010}.
Given the logical and linguistic properties of CNL 
(e.g. reasoning, paraphrasability, human- and machine- readability) the author
stresses that a successful mapping between NL and CNL can enable language based cognition of
simple autonomous software assistants, for reasoning 
and as interface to both peers and humans.

\section{Concept}
Given such rationale, the community should formulate a methodology for
operating a reduction of sentence-level natural language discourse, to 
a discourse representation formulated in a target controlled natural
language.

The author presents a possible pipeline abstraction 
of preexisting state-of-the-art means, as described in Figure 1. 
In particular, source channel text normalization (C1) to 
regularize erroneous phonetic transcriptions and spelling; 
a text to grounded Discourse Representation Structures (DRS)
parser (C2) which works thanks to Combinatory Categorial Grammar (CCG), i.e. 
a grammar formalism that allows a computationally efficient interface between 
syntax and structural semantics \cite{curran2007linguistically}.
The implemented form has already achieved optimal results and can produce Discourse Representation Structures as output \cite{Bos2008STEP2};
a previously trained sentence-level Support Vector Machine
(SVM) rule classifier,
which identifies the types of NL to CNL reductions that should be
operated (C3). A similarly implemented classifier is present in literature \cite{naughton2010sentence}.
We then have a syntactic manipulation engine to
transform the natural language input DRS into a set of compliant CNL DRS instances (C4), subject to the 
previously obtained classification results.
Such classification (C3) should account for, for instance:
\begin{itemize}
  \item intrinsically ambiguous natural language syntactic
  constructs 
  \item ambiguous anaphoric reference resolution
  \item conscious constraining decisions on the expressiveness 
  of specific CNL constructs 
\end{itemize}
The full enumeration of reduction case reasons is application 
domain-dependent and require an aprioristic study that can be
performed online and in a supervised manner, for instance with
active learning techniques.
A possible target CNL which has proven 
robustness and reliability is ACE \cite{fuchs:flairs2006}, which
has DRS to CNL verbalization functionalities, as well as
paraphrasing, proving and inference reasoning capabilities.
Figure \ref{drs} shows a simple instance of the presented pipeline,
which requires manipulation via sostitution of the unigram "linguistics" with the trigram "a linguistic class".

\begin{figure}[H]
\label{drs}
\begin{center}
{\sffamily \small
\color{Bittersweet} NL: "Harris can teach linguistics on Tuesdays."\\
{\huge $\Downarrow$}\\
\centering
{\sffamily \small \color{Bittersweet} ACE: "Harris can teach a linguistic class on Tuesday."}\\
}
\end{center}
\caption{Example of an NL sentence instance and a possible semantic-preserving reduction to ACE}
\end{figure}

\paragraph{Evaluation} Evaluation should mainly assess,
via the use of human evaluation, if given an arbitrary sentence 
related to the application domain, the meaning of this has been
successfully  conveyed to the target controlled sentence. 
For instance, a threshold of satisfactory quality in action-oriented tasking domains
\cite{nyga12actioncore} can be if arguments of intra-, mono-, di-
transitive verb arguments have been preserved, together with
correct anaphoric resolution. Evaluation will also assess domain-specific classification rates and computational efficiency.

\paragraph{Limitations}
The presented architecture does not make assumptions on the content of the predicates that are represented by words,
given that the manipulation is operated only at a structural level,
i.e. within the boundaries of DRS expressiveness. For a deeper 
predicate-related alignment, further considerations regarding lexicon
should be made, to provide word sense and Part-Of-Speech (POS) mappings between source 
vocabulary and target controlled vocabulary.

\paragraph{Potential}
Current statistic-based web search approaches that make use of word n-gram models can
exploit a more structural, discourse oriented
approach. Formalization enables logic 
satisfiability check of manipulated NL questions via reduction
and reasoning on First Order Logic (FOL) clauses.
The expressiveness of the latter would also allow reasoning as Constraint Satisfaction Problems (CSP), i.e. a widely adopted mathematical formalism that expresses real-world decision problems as unary and binary constraints over finite variable domains.
To pursue the example in Figure \ref{drs}, admitting other ontological knowledge of 
lecturers' availability and ability, we could formulate an NL question (that becomes a formal ACE question) to ask for
solutions to a simple timetable scheduling CSP problem, where the domains are the possible lecture days and types,
and the constraints are the required lecture types and time precedence relations between them.

\section{Future Work and Conclusions}
This concept-only presentation hopes to have briefly 
highlighted the potential that such abstract CNL-based
architecture can have, above all within the context of 
artificial assistants, as a means of interface, logic and combinatorial 
problem reasoning in ontology-based applications.
If compliant with CNL rules, a specific set of syntactically reduced NL
statements can seamlessly interface humans and machines while maintaining 
intelligibility and logical properties, such as entailment verification 
and inference. Future work should focus on implementation and efficiency verification 
of the stated architecture, to then investigate predicate-level 
(lexical) semantic alignment, to step towards (quasi-) complete 
sentence-level natural language formalization. 

\bibliographystyle{acl}

\bibliography{acl2013}

\end{document}